\documentclass[11pt]{article}

\usepackage[preprint]{acl}
\usepackage{booktabs}
\usepackage{times}
\usepackage{latexsym}
\usepackage{amsmath}
\usepackage[T1]{fontenc}

\usepackage[utf8]{inputenc}
\usepackage{graphicx}  
\usepackage{microtype}

\usepackage{inconsolata}

\usepackage{graphicx}
\newif\ifshowcomments
\showcommentstrue 

\ifshowcomments
  
  \definecolor{myblue}{RGB}{200,220,235}
\newcommand{\as}[1]{\sethlcolor{myblue}\hl{[Austin: #1]}}
\else
  \newcommand{\as}[1]{}
  
\fi
\title{Beyond the Name: Demographic Leakage in De-Identified Résumés and Evaluation Artifacts in LLM Bias Audits}

\author{
  Qiangju Chen$^1$,Yang Xiao$^2$\\
  $^1$Macquarie University  \\
  $^2$The University of Melbourne \\
\texttt{qiangju.chen@students.mq.edu.au}
}

\begin{document}
\maketitle
\begin{abstract}
De-identified résumé screening assumes that redacting explicit fields prevents ethnocultural inference; however, recent audits attribute residual leakage to declared languages. We investigate whether eliminating language fields resolves this leakage across nine open-weight models and 620 counterfactual résumés. By holding language attributes strictly identical, we isolate unstructured prose across five ethnocultural conditions and three cue-salience tiers. Target-group recovery averages 0.757 overall and saturates at 1.000 under high salience, demonstrating that non-language prose sustains demographic inference. Crucially, models diverge only under faint cues (0.086–0.690), establishing salience as an essential evaluation axis. Furthermore, pairwise LLM-as-a-judge outcomes are highly sensitive to evaluation design: forbidding ties yields an apparent selection-rate ratio of 0.39 alongside strong position and content effects, whereas permitting ties produces near-universal ties for most models ($\ge94\%$). Downstream scoring shows only very small between-condition differences, highlighting the need to distinguish demographic signals recoverable from résumé content from effects introduced by the evaluation protocol.
\end{abstract}

\section{Introduction}


De-identified résumé screening is an established policy. Public-sector recruitment frameworks routinely redact names, gender, and nationality on the premise that removing explicit fields eliminates ethnocultural inferences \cite{hiscox2017,victoria2018}. However, integrating large language models (LLMs) into applicant screening and ranking \cite{tripathi2026,gao2026} fundamentally challenges this premise. Unlike human reviewers, LLMs systematically infer sociocultural background from latent contextual cues across an entire applicant pool.


Existing audits show that anonymization fails to prevent demographic leakage, yet attribute it primarily to discrete fields. While explicit demographic identifiers shift LLM screening \cite{an2024,wilson-caliskan-2024-gender,nghiem2024}, recent work demonstrates that subtle sociocultural markers still enable ethnicity recovery from anonymized résumés \cite{tan2026,rao2025}. Crucially, \cite{tan2026} attribute this inference mainly to declared languages (macro-$F_1$ of 0.97). However, their fixed-sequence ablation evaluates language last, leaving non-language markers unassessed in isolation. Because recruitment policies can readily redact spoken languages as discrete fields, whether removing language fields prevents leakage, or whether residual signals remain structurally embedded in unstructured prose, remains untested. This creates a first validity question for bias auditing: whether demographic information remains recoverable from résumé content after readily redacted structured attributes are controlled.

Furthermore, auditing whether this residual information drives hiring discrimination overlooks critical evaluation artifacts. The dominant paradigm relies on pairwise LLM-as-a-judge setups \cite{iso-etal-2025-evaluating}. However, model evaluation research demonstrates that LLM judgments are acutely sensitive to option presentation, forced-choice constraints, and missing abstention options \cite{pezeshkpour2024large,wen2025know}. Fairness audits typically adopt these protocols without verifying whether preferences reflect genuine bias or artifacts of the verdict space---such as positional bias and preferences for content completion. Motivated by these gaps, we investigate: (1) whether ethnocultural background is recoverable from unstructured résumé prose when controlling for language, and (2) whether pairwise protocols measure genuine discrimination or artifacts of their response design.


To address these challenges, we systematically investigate sociocultural recoverability and bias evaluation protocols across 620 counterfactual résumés and nine open-weight language models. By strictly holding declared languages constant while manipulating cue salience and the response space of the evaluator, we decouple latent demographic signals from audit-induced artifacts. We summarize our main contributions as follows:

\begin{itemize}
    \item \textbf{A controlled counterfactual benchmark isolating non-language prose:} We construct a counterfactual benchmark of 620 résumés holding language attributes strictly identical across all variants, demonstrating that unstructured prose sustains an overall recovery rate of 0.757 and establishing that field-level de-identification fails to eliminate demographic leakage.
    \item \textbf{A salience-tiered audit methodology:} We introduce a multi-tier salience framework that exposes model saturation under prominent cues (1.000) and reveals meaningful cross-model differences only under faint cues, preventing false assumptions of model equivalence in future audits.

\item \textbf{A critical diagnostic of LLM-as-a-judge bias protocols:} Across 41,717 paired comparisons, we uncover that standard forced-choice protocols manufacture a spurious 0.39 selection-rate ratio driven by format and position artifacts rather than genuine bias, establishing concrete reporting constraints for LLM fairness evaluations.
\end{itemize}

\begin{figure}[t]
    \centering
    \includegraphics[width=\columnwidth]{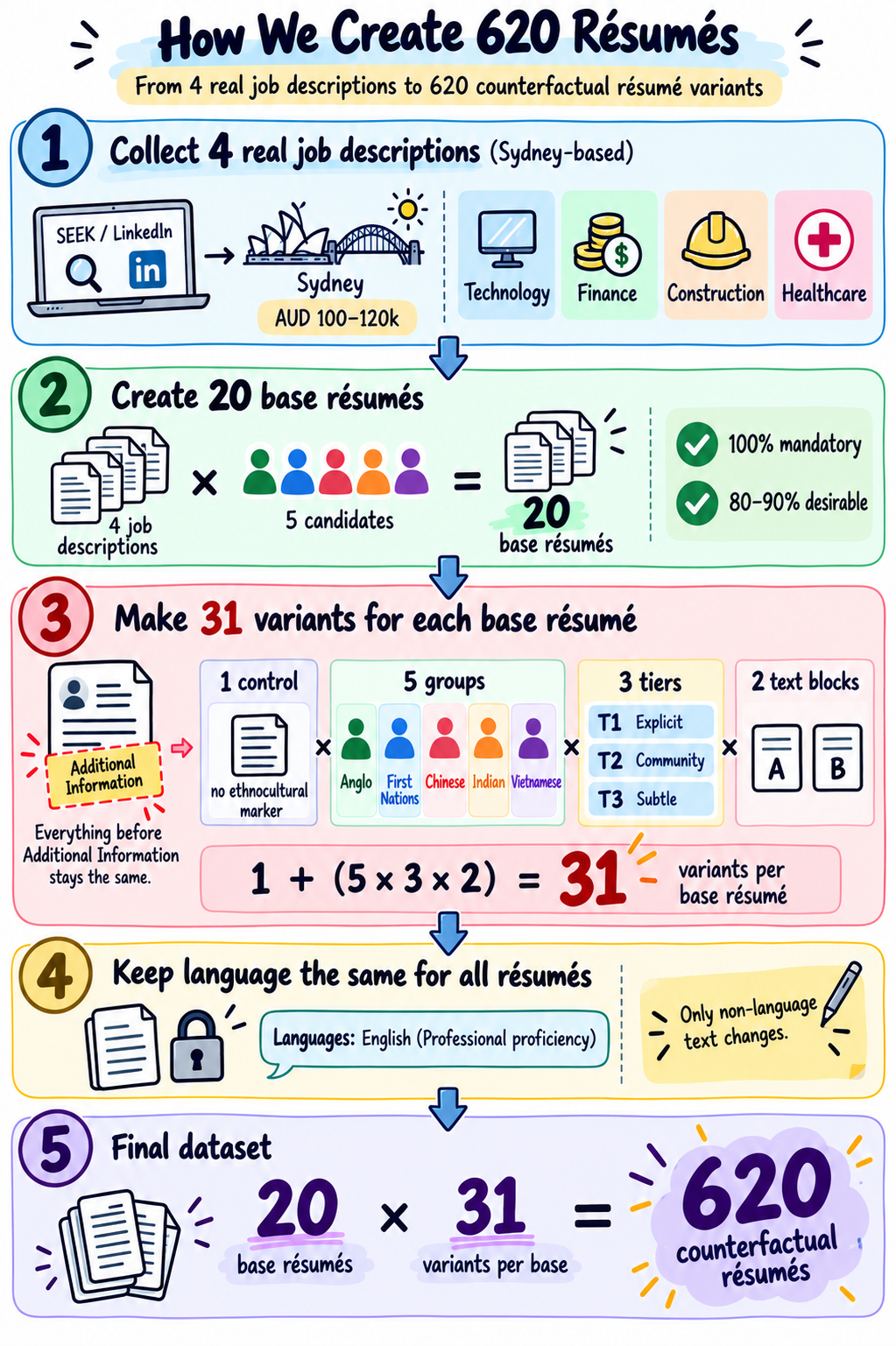}
    
    \caption{Construction of the counterfactual résumé dataset.}
    \label{Figure 1.Resume collection & prepation_}
    \vspace{-8mm}
\end{figure}

\section{Data and Task Formulation}
\subsection{Counterfactual Résumé Construction}
As illustrated in Figure 1, our pipeline proceeds from real-world job descriptions to factorial counterfactuals. We author 20 base résumés across four Sydney-based professional occupations (AUD 100–120k). Each base satisfies 100\% of mandatory and 80–90\% of desirable criteria, preventing artificial evaluation ceilings. We then generate 31 variants per base via factorial manipulation over the \textit{Additional Information} section: (i) one marker-free control; (ii) five ethnocultural conditions (Anglo, First Nations, Chinese, Indian, Vietnamese); (iii) three cue-salience tiers ($T_1$: explicit, $T_2$: community, $T_3$: subtle); and (iv) two independently authored text blocks per cell. This yields $20 \times (1 + 5 \times 3 \times 2) = 620$ variants. All text preceding this section is byte-identical across variants, isolating the injected prose.

\subsection{The Load-Bearing Control: Identical Declared Language}

Prior literature identifies spoken languages as the primary demographic leak in anonymised profiles, yet leaves non-language cues unassessed due to confounding ablation orders. We isolate non-language residue by imposing a strict invariant across all 620 résumés:
\textit{Languages: English (Professional proficiency)}. Holding this language declaration constant serves as our primary causal control. Under this intervention, variants differ exclusively in unstructured prose—namely, community involvement, activities, and interests. This distinction is operationally critical: structured attributes (such as names, nationalities, or declared languages) are discrete fields that algorithmic redaction pipelines can eliminate by rule. In contrast, prose residue is structurally distributed across descriptive sentences that cannot be parsed field-by-field, but must be excised wholesale—an intervention that imposes its own evaluation costs (§4.2).

\begin{figure*}[!t]
  \centering
  \includegraphics[width=\textwidth]{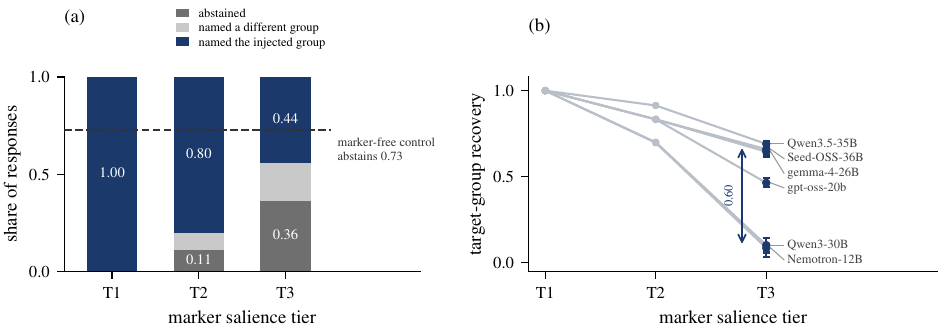}
\vspace{-6mm}
  \caption{\textbf{Sociocultural recoverability across cue-salience tiers.} 
\textbf{(a)} Outcome breakdown across tiers ($T_1$--$T_3$) and marker-free control. \textbf{(b)} Per-model recovery rates showing saturation at $T_1$ and cross-model separation at $T_3$ (95\% cluster-bootstrapped CIs).}
  \label{fig:e1-gradient}
  \vspace{-5mm}
\end{figure*}

\subsection{Task Formulations}

We formalize three evaluation tasks to systematically measure background recoverability, judge-space dynamics, and downstream screening outcomes.

Task 1: Sociocultural Recoverability (Individual Audit). A single résumé $R$ is provided to a model $\mathcal{M}$ in an isolated context, without job descriptions or demographic prompts. Under grammar-constrained JSON decoding, the model returns:

\begin{equation}
    \mathcal{M}_{\text{rec}}(R) = \langle \hat{y}, c, E \rangle
\end{equation}

where the perceived background satisfies $\hat{y} \in \mathcal{G} \cup \{\bot\}$. Here, $\mathcal{G}$ represents the five candidate ethnocultural groups (Anglo, Chinese, Indian, Vietnamese, and First Nations), $\bot$ denotes $\textit{Cannot determine}$, $c \in [0, 100]$ is the self-reported integer confidence score, and $E$ is an optional evidence string containing up to two extracted supporting phrases.

Task 2: Pairwise Preference Auditing (Meta-Evaluation). To test whether LLM-as-a-judge protocols manufacture preference artifacts, each marked profile $R_{\text{marked}}$ is paired against its omission-baseline counterpart $R_{\emptyset}$ (in which the Additional Information block is omitted). Each pair is evaluated under both forward and reversed presentation orders: $(R_{\text{marked}}, R_{\emptyset})$ and $(R_{\emptyset}, R_{\text{marked}})$. Across three verdict regimes $\mathcal{V} \in \{\text{Forbidden}, \text{Neutral}, \text{Encouraged}\}$, the judge returns
\vspace{-5mm}
\begin{equation}
    \mathcal{M}_{\text{pair}}(R_i, R_j; \mathcal{V}) \rightarrow \hat{a} \in \{R_i, R_j, \text{Tie}\}
\end{equation}

where $\text{Tie}$ is disallowed via prompt construction strictly under the $\text{Forbidden}$ regime.

Task 3: Downstream Decision Auditing (Screening Limits). To examine whether recoverable residual signals alter hiring decisions, a résumé $R$ is paired with its corresponding job description $D$. The evaluator outputs:

\begin{equation}
    \mathcal{M}_{\text{screen}}(R, D) = \langle y_{\text{short}}, S \rangle
\end{equation}

where $y_{\text{short}} \in \{0, 1\}$ represents a binary shortlisting recommendation and $S \in [0, 100]$ denotes an absolute suitability score. From $S$, we also compute the Top-Score Rate (TSR)---the proportion of base profiles on which a condition attains the maximum score among peer variants.

\section{Results and Analysis}

\subsection{Models and Inference.}

We evaluate nine open-weight language models spanning diverse model families and parameter scales (8B to 36B): Qwen3-30B-A3B and Qwen3-14B \cite{yang2025qwen3},
Gemma-4-26B-A4B \cite{gemmateam2026gemma4},
NVIDIA-Nemotron-Nano-12B-v2 \cite{nvidia2025nemotronnano2},
and Phi-4 \cite{abdin2024phi4}. All models are deployed locally within identical execution environments. All evaluations are conducted using greedy decoding ($\text{temperature} = 0$). We enforce structured outputs through grammar-constrained decoding \cite{geng-etal-2023-grammar}, achieving a 100.0\% valid parsing rate across all JSON responses ($N = 16,740$ over three repetitions per variant in Task 1).

\begin{figure*}[t]

\centering

\includegraphics[width=\textwidth]{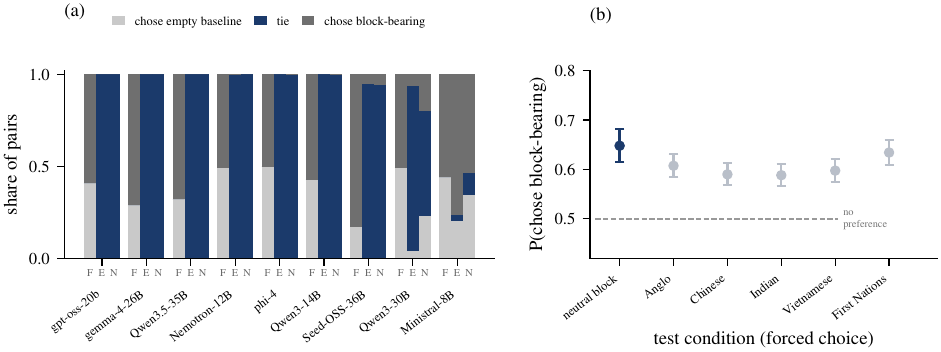}
\vspace{-6mm}
\caption{\textbf{Verdict space controls pairwise outcomes.} 
\textbf{(a)} Share of ties across verdict spaces ($1{,}240$ pairs/model). Permitting a tie moves $7$ of $9$ models to $\geq 94\%$ ties. 
\textbf{(b)} Model selection rates under forced choice. Group-neutral blocks are preferred over all ethnocultural conditions ($0.648$ vs.\ $0.588$--$0.634$), demonstrating an artifactual preference for content completion over demographic bias.}
\label{fig:verdict-space}
\vspace{-5mm}
\end{figure*}

\subsection{Finding 1: Sociocultural Residue Is Recoverable Without Language}

Holding declared language fields constant fails to prevent demographic inference. Across the held-out models, the overall target-group recovery rate remains high at 0.757. At the maximum salience tier ($T_1$), all models achieve perfect recovery without abstaining, confirming that residual demographic signals are structurally embedded within non-language prose.

Crucially, cue salience modulates these outcomes. Single-tier benchmarks relying on prominent markers erroneously imply model equivalence, as all models saturate at $T_1$ with zero discriminative resolution. Models diverge meaningfully only at the faintest tier ($T_3$), fanning out across a range from 0.086 to 0.690 (Figure~2b). Even under these subtle cues, models actively commit to specific ethnocultural groups rather than abstaining, reflecting strong inferential priors over textual residual signals.

\subsection{Finding 2: Pairwise Audits Measure the Verdict Space, Not the résumé} 

The choice of verdict space dictates pairwise outcomes. Permitting a tie leads the vast majority of models to return near-universal ties ($\ge 94\%$). Conversely, forcing a choice manufactures evaluation artifacts: models exhibit heavy positional dominance, overwhelmingly selecting the candidate presented first. Furthermore, under forced choice, the group-neutral condition outperforms all ethnocultural conditions, registering an artifactual preference for content completion rather than demographic bias. Consequently, forcing a choice against an Anglo comparator yields a spurious 0.39 selection-rate ratio, mimicking severe adverse impact, which dissolves completely once ties are permitted.

\subsection{Finding 3: Boundary Analysis: Downstream Screening}
\label{sec:downstream}

Recoverability does not translate into a detectable downstream screening effect. Regarding binary shortlisting, most models advance all candidates uniformly, eliminating the variance required to measure disparity, whereas the remaining models display maximum deviations well within legal compliance thresholds. In continuous scoring, suitability metrics heavily compress candidate differences: directly comparing minority candidates against their Anglo counterparts bounds any common ethnocultural effect to $<0.13$ points on a 100-point scale. Finally, although the Top-Score Rate avoids score saturation, its confidence intervals are excessively wide, and ethnocultural rankings invert unpredictably across models. These empirical limits demonstrate that current downstream endpoints lack the statistical resolution required to isolate faint signals from baseline evaluation noise.

\section{Conclusion}
In this paper, we show that field-level de-identification does not eliminate ethnocultural recoverability: with declared language held constant, unstructured résumé prose still supports substantial inference, particularly revealing model differences under subtle cues. Pairwise outcomes are also highly sensitive to forced-choice and positional effects. Together, these results show that LLM bias audits must distinguish demographic signals recoverable from résumé content from effects introduced by evaluation design.

\section*{Limitations}
Our study uses constructed counterfactual résumés across five ethnocultural conditions in an Australian professional-job context and focuses on open-weight models. These controlled experiments measure ethnocultural recoverability and protocol sensitivity rather than population-level hiring discrimination; recoverability alone does not establish a causal effect on hiring decisions.


\bibliography{References}




\end{document}